\documentclass{article}
\usepackage{spconf,amsmath,graphicx}
\usepackage{times}
\usepackage{epsfig}
\usepackage{graphicx}
\usepackage{xspace}

\usepackage{amsmath}
\usepackage{amssymb}

\usepackage{subcaption}
\usepackage{bm}                   
\usepackage{multirow}
\usepackage{amsmath}
\usepackage[table,xcdraw]{xcolor}
\usepackage{numprint}
\usepackage[export]{adjustbox}
\makeatletter
\DeclareRobustCommand\onedot{\futurelet\@let@token\@onedot}
\def\@onedot{\ifx\@let@token.\else.\null\fi\xspace}

\newif\ifwithlatentalgo
\newif\iflong
\newif\iflettrine
\newif\ifiksvm
\newif\ifaptable
\longtrue
\newif\ifmultitables
\multitablestrue
\newif\ifappendix
\appendixtrue
\newif\iflongtables
\newif\ifltpap
\newif\ifnooverview

\def\psm{PSMNet\xspace}
\def\ganet{GANet\xspace}
\def\ganete{GANet11\xspace}
\def\ganetd{GANetdeep\xspace}
\def\mnet{MobileNet\xspace}

\def\gmacs{GMACs\xspace}
\def\macs{MACs\xspace}

\def\kittif{KITTI\xspace2015\xspace}
\def\sf{SceneFlow\xspace}

\def\sns{stereo networks\xspace}

\def\td{3D\xspace}

\def\tdcs{3D convolutions\xspace}

\def\stdc{separable \td convolution\xspace}
\def\stdcs{separable \td convolutions\xspace}

\def\FWSC{FwSC\xspace}
\def\FWSCs{FwSCs\xspace}

\def\DWSCs{DwSCs\xspace}

\def\FDWSCs{FDwSCs\xspace}

\def\ie{\emph{i.e}\onedot}

\def\wrt{\emph{w.r.t}\onedot}

\def\cf{\emph{c.f}\onedot}

\def\etc{\emph{etc}\onedot}

\def\vs{\emph{vs}\onedot}

\def\eg{\emph{e.g}\onedot}

\def\para#1{{\bf #1.}}

\def\mbold#1{\textbf{#1}}
\def\mitalic#1{\textit{#1}}

\newif\ifwithcomments 
\withcommentsfalse
\ifwithcomments

\long\def\comment#1{\para{\sffamily\{***\color{RoyalBlue}#1}****\} }%

\long\def\discuss#1{{\color{ForestGreen}\mbold{\underline{DISCUSS:}}#1} }
\long\def\diagrams#1{\\{\color{Red}\mbold{\underline{Diagram:}}#1} }

\long\def\idea#1{{\\\color{Magenta} \mbold{Idea:}#1}}
\long\def\revcomment#1{Reviewer BMVC:#1 }
\else

\long\def\comment#1{}
\long\def\discuss#1{ }
\long\def\revcomment#1{ }
\long\def\diagrams#1{ }
\long\def\idea#1{}
\fi

\def\l2norm{\mitalic{L2Norm}}

\def\figref#1{Fig.~\ref{fig:#1}}

\def\tabref#1{Table~\ref{table:#1}}
\def\secref#1{Sec.~\ref{sec:#1}}

\def\eqref#1{Eq.~(\ref{eq:#1})}

\def\vs{\ensuremath{\vec{s}}\xspace}

\newcounter{rno}

\usepackage[pagebackref=true,breaklinks=false,colorlinks,bookmarks=false]{hyperref}

\def\vs#1{\vspace*{#1}}
\def\ivs#1{}
\def\params{Params (M)\xspace}
\def\ops{Ops (\gmacs)\xspace}

\title{Separable convolutions for optimizing \td Stereo Networks\ivs{-10pt}}
\name{Rafia Rahim, Faranak Shamsafar and Andreas Zell \ivs{-10pt} \thanks{This work is part of the project DeepStereoVision (FRE: 01IS18024B) sponsored by the German Ministry of Education \& Research (BMBF).}
	\vspace{-14pt}
}

\address{	\tt \small \{rafia.rahim, faranak.shamsafar, andreas.zell\}@uni-tuebingen.de\\ \small Department of Computer Science (WSI), University of Tuebingen, Germany}

\begin{document}
%
\maketitle
\begin{abstract}
Deep learning based 3D stereo networks give superior performance compared to 2D networks
and conventional stereo methods. However, this improvement in the performance comes at the cost 
of increased computational complexity, thus making these networks non-practical for the real-world applications.
Specifically, these  networks use 3D convolutions as a major work horse to refine and regress disparities. In this work first, we show that  these 3D convolutions in \sns consume up to $94$\%  of overall network operations and  act as a major bottleneck. Next, we propose a set of ``plug-\&-run" separable convolutions to reduce the number of parameters and operations. When integrated with the existing state of the art \sns, these convolutions lead up to $7\times$ reduction in number of operations and up to $3.5 \times$ reduction in parameters without compromising their performance. In fact these convolutions lead to improvement in their performance in the majority of cases\footnote{\href{https://github.com/cogsys-tuebingen/separable-3D-convs-for-stereo-matching}{Code link}.}.
\end{abstract}
%
\vs{-5pt}
\section{Introduction}
\ivs{-3pt}
Estimating depth from an input stereo pair is an actively pursued goal in the computer vision domain. This typically involves two steps: (i) image rectification based on epipolar constraints and (ii) disparity estimation. 
A typical disparity estimation pipeline mainly consists of matching cost computation and cost aggregation steps~\cite{scharstein2002taxonomy}. In the matching cost step, features are computed from both images and matched  to generate a potential set of matches. In the cost aggregation step, global  \cite{hosni2012fast, yoon2006adaptive}  or local neighbourhood \cite{min2011revisit, scharstein2002taxonomy} constraints are applied to refine the potential matches and generate the disparity values.
\begin{figure}[!t]
\centering
\includegraphics[width=0.5\textwidth]{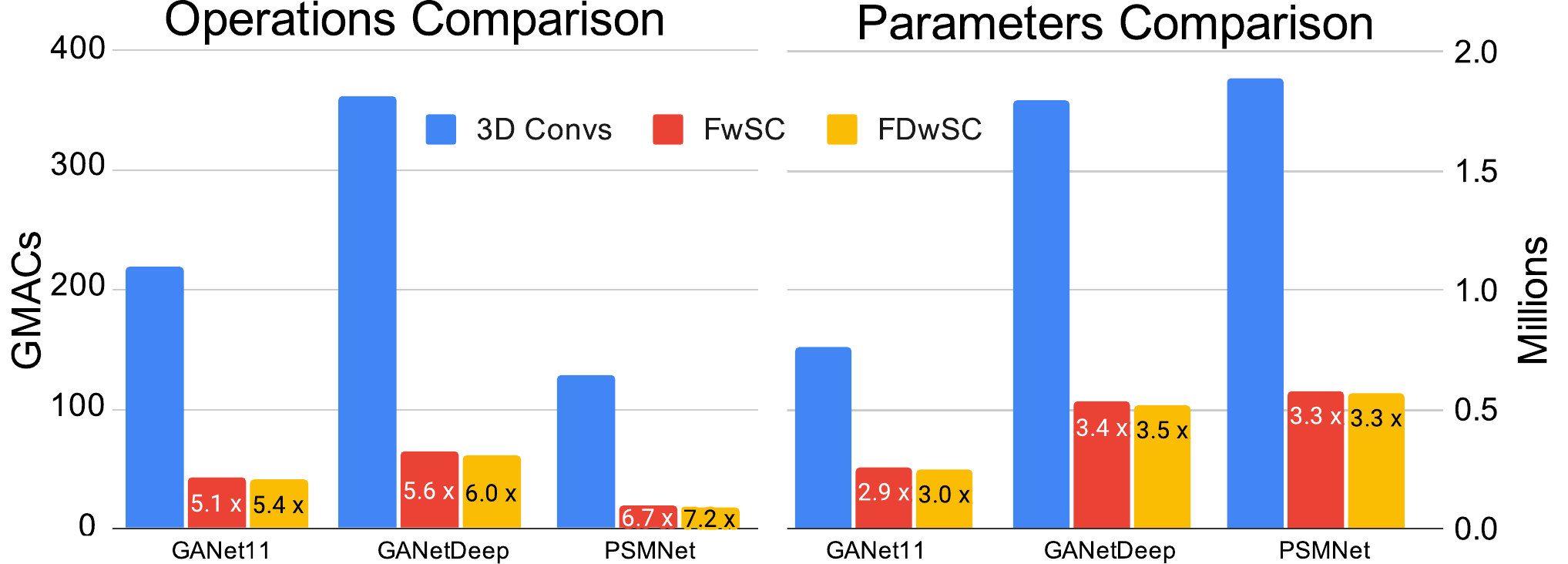}
\caption{Relative comparison between \tdcs, \FWSCs, \FDWSCs \wrt number of operations and parameters. Here $\times$ represents the reduction factor.}
\label{fig:3dsc-comp}
\vs{-15pt}
\end{figure}

With the renaissance of deep learning, many new methods have been proposed to improve the overall density estimation pipeline. These methods target either a specific step \cite{zbontar_stereo_2016,seki_sgm-nets:_2017,luo2016efficient} or work in end-to-end fashion \cite{kendall2017endtoend,chang_pyramid_2018,zhang_ga-net:_2019} to improve the overall pipeline.  End-to-end deep learning methods, in general, have two main stages. In the first stage, a backbone network is used to compute features from both the input images. In the second stage, features from both the images are merged to form a cost volume; this cost volume is then further processed via a sequence of convolutional layers to regress the final disparities. Broadly speaking, these end-to-end methods can be classified to 2D or \td methods -- depending on their methodology of merging features, types of layers and convolutions in the second stage -- \cf \secref{related-work}.  Overall, 2D networks use 2D convolutions after merging features, whereas 3D methods use 3D convolutions over 4D cost volumes to regress and refine disparities.

Even though \td methods give far superior performance compared to 2D methods,  this performance  improvement comes at the cost of extra computational budget. In fact, these \td methods are orders of magnitude slower than 2D ones. A lot of efforts in the community have been recently targeted to  make these methods computationally efficient \cite{wang_anytime_2019,tulyakov2018practical,duggal_deeppruner:_2019}. Although majority of these methods target optimization of the cost-volume construction step, they still use the costly \tdcs for disparity regression. In this work, we aim to further reduce the overall computational footprint of the \td methods by optimizing \tdcs.

For this, we first do a thorough empirical investigation of the state of the art \td stereo methods \cite{zhang_ga-net:_2019, chang_pyramid_2018} to identify major contributing factors to the overall computational cost of these methods. As expected, our investigations conclude that \tdcs are by far the most costly operations in these methods. For instance, in \ganetd, \tdcs consume around 94\% of operations -- \cf \tabref{cost}. Next, we propose methods for optimizing \tdcs, \ie how we can optimize \tdcs without sacrificing \td \sns performance. 

To this end, motivated by the recent performance of light-weight \td networks for visual recognition \cite{qiu2017learning, ye_3d_2018}, we propose to replace costly \td convolution operations in \td \sns with their light-weight counterparts (\ie \stdcs) to build efficient \sns. For this we design and explore three different versions of \stdcs: (i) Feature-wise Separable Convolutions (\FWSCs);  (ii) Disparity-wise Separable Convolutions (\DWSCs); and their extremely separable version (iii)  Feature-\&-Disparity-wise Separable Convolutions (\FDWSCs).  We empirically evaluate these \stdc versions as  ``plug-\&-run" replacement for existing \tdcs. 

Overall, we make the following contributions:  (i) we present an in-depth cost
analysis of major building blocks of state of the art \td \sns; (ii) we design
and propose separable \tdcs for \sns; (iii) we
empirically evaluate and show the ``plug-\&-run" characteristic of proposed
convolutions for state of the art methods on standard datasets. Our results show that \FWSCs and \FDWSCs can be reliably used  as a replacement for \tdcs in \sns without sacrificing their performance while reducing the number of operations (up to $7.2\times$) and number of parameters (up to $3.5\times$) -- \cf \figref{3dsc-comp} \&  \tabref{results}.  To the best of our knowledge, this is the first work to explicitly optimize \td convolutions involved in  \td \sns. 
\vs{-5pt}
\section{Related Work}\label{sec:related-work}
Generally speaking, deep learning stereo methods can be broadly categorized into 2D or 3D learning methods~\cite{poggi_synergies_2020} based on how they merge information across views and then process it. For instance, 2D methods merge features from the left and right  images into a 3D cost volume and then apply 2D convolutions. DispNet-C \cite{mayer_large_2016}, iResNet \cite{liang_learning_2018}, MADNet \cite{tonioni_real-time_2018}, SegStereo \cite{yang_segstereo_2018}, EdgeStereo \cite{song_edgestereo_2018} are some examples of 2D methods. On the other hand, \td methods first generate a 4D cost volume by merging features from the left and right images (under stereo constraints) and use 3D convolutional layers to refine and regress dense disparities. GC-Net \cite{kendall2017endtoend}, \psm \cite{chang_pyramid_2018}, \ganet \cite{zhang_ga-net:_2019}, AnyNet \cite{wang_anytime_2019} and DeepPruner \cite{duggal_deeppruner:_2019} are some examples of 3D methods. Overall 3D methods produce better results but at the expense of extra computational cost. Although recent methods propose to optimize the 3D cost volume construction step,  they still use costly 3D convolutions for cost aggregations, refinement and disparity regression. For instance, \cite{wang_anytime_2019} use hierarchical \td cost volumes  to reduce  4D volume construction cost while \cite{tulyakov2018practical} introduce bottleneck matching modules to efficiently match features across disparities. In another work \cite{duggal_deeppruner:_2019} employs patch-matching to sample a range of disparities during volume construction.  
\vs{-5pt}
\section{Methodology}
\label{sec:Methodology}
Generally, an end-to-end \td stereo pipeline consists of these modules: (i) Feature extraction: a 2D shared backbone-network to extract features from left and right images of a rectified stereo pair; (ii) Volume construction: a 4D cost volume constructed by merging left and right images features maps; and (iii) Cost aggregation \& disparity regression: a \td network to aggregate cost volume and then regress and refine disparities from aggregated cost volume.
\begin{table}[!t]
\centering
\let\footnoterule\relax
\renewcommand{\arraystretch}{1.25}
	\caption{Detailed computational cost profiling of \td \sns.}
	\vspace{-8pt}
\resizebox{0.47\textwidth}{!}{\begin{minipage}{0.7\textwidth}
	\label{table:cost} 
	\npdecimalsign{.}
	\nprounddigits{1}
	\begin{tabular}{|l|l|n{2}{1}|n{2}{1}|n{3}{1}|n{2}{1}|n{3}{1}|}
		\hline
		\multirow{2}{*}{Model} 
		&
		\multirow{2}{*}{Metric} &
		\multicolumn{2}{l|}{Feature Network} &
		\multicolumn{2}{l|}{\begin{tabular}[c]{@{}l@{}}Cost Aggregation and\\ Disparity Regression\end{tabular}} &
		Total \\
		\cline{3-7} 
		&                                                          & \multicolumn{2}{l|}{2D CNNs} & \multicolumn{2}{l|}{3D CNNs} & \\ \hline
		
		\multirow{2}{*}{\ganete}     & \# \params     & $2.37$      & $53.33\%$     & $0.76$       & $17.7\%$    & $4.48$             \\ 
		\cline{2-7} 
		&                                    \# \ops & $4.80$      & $2.03\%$      & $217.44$     & $91.92\%$    & $236.56$           \\ \hline				\multirow{2}{*}{\ganetd}   & \# \params     & $2.37$      & $36.31\%$     & $1.80$       & $27.21\%$    & $6.58$             \\ 
		\cline{2-7} 
		&                                   \# \ops & $4.80$      & $1.25\%$      & $358.85$     & $93.51\%$    & $383.75$           \\ \hline
		\multirow{2}{*}{\psm}        & \# \params     & $3.34$      & $63.52\%$     & $1.89$       & $36.5\%$     & $5.22$             \\ 
		\cline{2-7} 
		&                                    \# \ops & $58.19$     & $31.40\%$     & $127.12$     & $68.6\%$     & $185.31$           \\ \hline
	\end{tabular}
	\npnoround
\footnotetext{Note: Here $\%$ represents the fraction of parameters (Params in Millions) and operations (Ops in \gmacs) relative to complete network (image size for \ganet models is $240\times528$ and \psm is $256\times512$ pixels.).  We can see that  although 3D CNNs have fewer parameters compared to 2D CNNs, they still consume the majority of operations.}
	
\end{minipage}}
\vs{-16pt}
\end{table}
\vs{-5pt}
\subsection{Profiling \td Stereo Networks}
\vs{-1pt}
\label{sec:Profiling}
As the first step, we profile the overall computational requirements of the state of the art \td \sns \cite{zhang_ga-net:_2019,chang_pyramid_2018}. Specifically, we record the number of parameters and operations involved in feature extraction and cost aggregation \& disparity regression modules -- \cf \tabref{cost}. Here operations are reported as the number of \macs (Multiply-ACcumulate operations) where $ 1  \mbox{MAC} =  1 $ multiplication + $ 1 $ addition operations. We can observe from \tabref{cost} that despite having fewer parameters than 2D networks, 3D networks consume more operations in all the networks and thus represent clear bottlenecks. For instance, in \ganete  \cite{zhang_ga-net:_2019} 3D CNN contains $0.8 \times 10^{6}$ ($17\%$ relative to complete network) parameters compared to 2D feature extraction module's $2.4 \times 10^{6}$ parameters ($52.8\%$). However, it consumes $217$ \gmacs  operations  ($91.9\%$ of overall operations) compared to $4.8$ \gmacs ($2$ \%) of the 2D network. Similar observations can be made about other networks.  

Once we have built a detailed insight into the reasons behind these networks computational overhead, we set out to network design choices to reduce the computational overhead without compromising their performance.
\vs{-5pt}
\subsection{Separable 3D Stereo Networks}
\vs{-1pt}
\begin{figure*}[!t]

\begin{minipage}[t]{0.58\textwidth}		
\begin{tabular}{l@{ }l}
				\hspace*{-10pt}\includegraphics[width=0.43\textwidth]{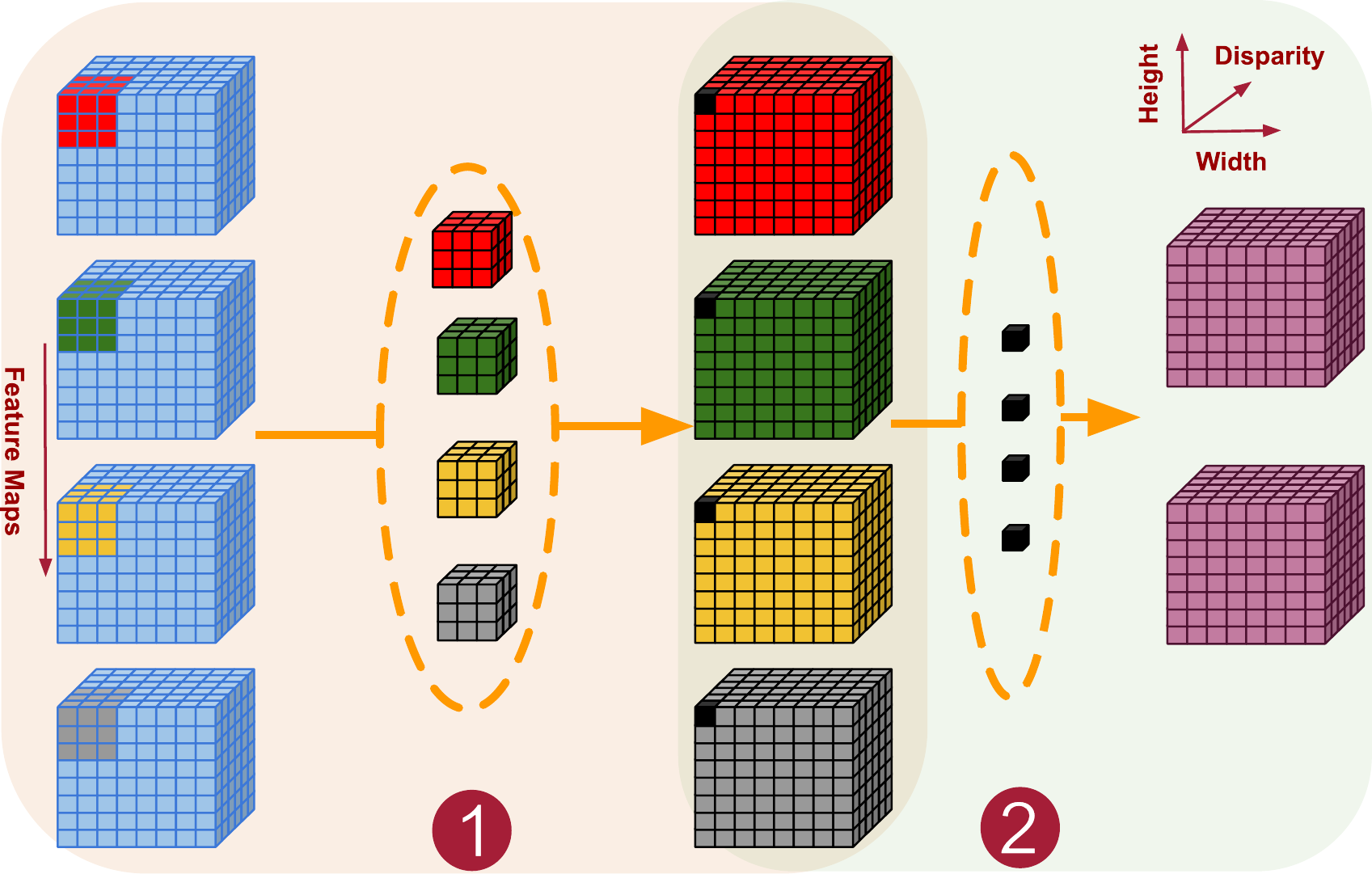}
		&  	\includegraphics[width=0.59\textwidth]{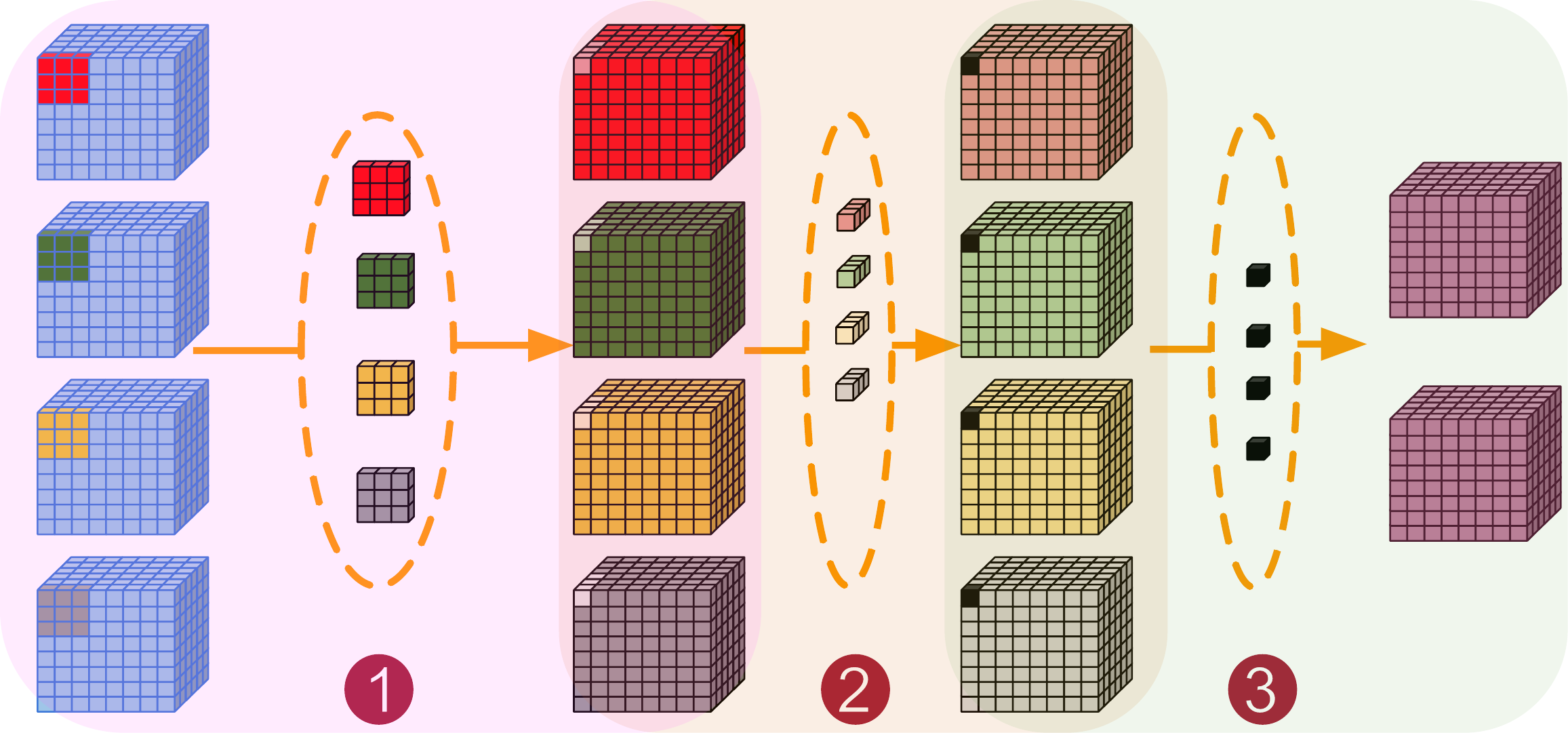}\\
	\end{tabular}
\vspace*{-8pt}
\captionof{figure}{Pictorial description of steps involved in the computation of \FWSCs (left) and \FDWSCs (right).}
\label{fig:3dsc}
\end{minipage}\hspace*{5pt}
\begin{minipage}[t]{0.4\textwidth}
\resizebox{0.47\textwidth}{!}{\begin{minipage}{0.92\textwidth}
	\renewcommand{\arraystretch}{1.5}
\npdecimalsign{.}
\nprounddigits{2}
\begin{tabular}{|l|l|l|n{3}{2}|n{2}{2}|c|n{2}{2}|c|}
\hline
\multirow{2}{*}{Model} & \#3D & \multirow{2}{*}{Metric} &\multicolumn{1}{c|}{3D}&\multicolumn{2}{c|}{\FWSCs} & \multicolumn{2}{c|}{\FDWSCs} \\
\cline{5-8}
& Convs & &Convs &&Reduction& & Reduction\\ 
\hline
\multirow{2}{*}{\ganete}   & \multirow{2}{*}{$11$} & \# \params      & $0.76$   & $0.26$  & $2.9 \times$ & $0.25$  & $3.0 \times$ \\ \cline{3-8} 
&                       & \# \ops & $217.44$ & $43.05$ & $5.1 \times$ & $40.60$ & $5.4 \times$ \\ \hline
\multirow{2}{*}{\ganetd} & \multirow{2}{*}{$15$} & \# \params      & $1.80$   & $0.53$  & $3.4 \times$ & $0.52$  & $3.5 \times$ \\ \cline{3-8} 
&                       & \# \ops & $358.85$ & $64.03$ & $5.6 \times$ & $60.06$ & $6.0 \times$ \\ \hline
\multirow{2}{*}{\psm}        & \multirow{2}{*}{$22$} & \# \params      & $1.89$   & $0.58$  & $3.3 \times$ & $0.57$  & $3.3 \times$ \\ \cline{3-8} 
&                       & \# \ops & $127.12$ & $19.06$ & $6.7 \times$ & $17.70$ & $7.2 \times$ \\ \hline
\end{tabular}
\npnoround
\end{minipage}
}
\vspace*{-4pt}
\captionof{table}{Comparison of \tdcs parameters and operations with separable \td Convs.}
\label{table:stats-convs}
\end{minipage}
\vs{-12pt}
\end{figure*}
\label{sec:separable-networks}
Recently light-weight convolutional networks (like \mnet, ShuffleNet, \etc \cite{sandler2018mobilenetv2,zhang2018shufflenet}) have been proposed to improve the computational efficiency of 2D visual recognition convolutional networks. These methods propose to use depth-wise separable variants of 2D convolutions to reduce the number of operations. However, in comparison to a 2D convolution kernel, a 3D convolution kernel works on a 4D volume and aggregates information across spatial, depth and channels dimensions, and thus there are different possible ways a 3D convolution can be made separable.  
For instance, a 3D kernel ($k \times k \times k$, \eg $k=5$) working on an input 4D cost volume of size  $h_i \times w_i \times d_i \times c_i$ (input height, width, depth (disparity), and channels) aggregates information from a window of size $k \times k \times k \times c_i$ in each of its applications. Furthermore, as a 3D convolutional layer contains many 3D kernels each producing a new feature map so overall there are $(h_{i} \times w_{i} \times d_{i} \times (k \times k \times k \times c_{i})) \times c_{o} $ operations performed in each layer -- here $c_o$ is number of output channels (number of kernels) in a layer. From this, we can see that these 3D convolutions  can be made separable across either $d_i$ or $c_i$ dimension or even simultaneously both across $d_i$ and $c_i$. We intend to reduce computational requirement by making \tdcs separable in one or more of the given dimensions.
\vs{-8pt}
\subsubsection{Feature-wise Separable Convolutions (\FWSCs):}
\vs{-2pt}
As a first replacement for \tdcs in \sns, we propose to split the convolutions across the feature (or channel) dimension. \FWSCs works in two steps: in the first step,  the 4D cost volume of size $h_i \times w_i \times d_i \times c_i$ is split into $c_i$ cubes, and $c_i$ kernels of size $ k \times k \times k$ are applied to generate $c_i$ output cubes -- \cf \figref{3dsc} (left). 

In the second step, $c_o$ point-wise kernels of size $1\times1\times1\times c_i$ are applied to aggregate information across the $c_i$ cubes. These steps lead to a significant reduction in the number of operations, as total the number of operations reduces to $(h_{i} \times w_{i} \times d_{i} \times k \times k \times k) \times c_{i}  +  (h_{i} \times w_{i} \times d_{i} \times 1 \times 1 \times 1 \times c_{i}) \times c_{o}$ compared to \tdcs. When applied in the state of the art \sns, \FWSCs lead up to $6.7\times$ reduction in \gmacs (see  \tabref{stats-convs})\footnote{Please note that these calculations also include \macs operations for biases and batch-normalization.} with better performance than \tdcs in the majority of cases (\secref{Results}).
\vs{-8pt}
\subsubsection{Disparity-wise Separable Convolutions (\DWSCs):}
\vs{-2pt}
\DWSCs are identical to \FWSCs, except the separable operation is performed in the disparity dimension. Precisely, in the first step the 4D cost volume of size $h_i \times w_i \times d_i \times c_i$ is split into $d_i$ cubes (after permuting $h_i \times w_i \times d_i \times c_i$ to $h_i \times w_i \times c_i \times d_i$) and $d_i$ kernels of size $ k \times k \times k$ are applied to produce $d_i$ output cubes.

In the second step, $c_o$ kernels of size $1\times1\times1\times d_i$ are applied to aggregate information across the $d_i$ cubes. Finally, the output
volume is permuted back. Although \DWSCs lead to a reduction in the number of operations, the overall reduction is not as significant as 
in \FWSCs, because usually in \sns size of disparity dimension is greater than the size of feature dimension.  For instance, in \ganetd
the size of disparity dimension is 48 whereas in almost all the layers feature dimension is 32.  

Moreover, for a given pixel first aggregating the information across the feature dimension and then across the disparity dimension does not appear to be the optimal choice. This was confirmed by our initial experiments, which returned much worse results thus we do not report further experimental results on the \DWSCs.
\vs{-8pt}
\subsubsection{Feature-Disparity-wise Separable Convolutions (\FDWSCs)}
\vs{-2pt}
\FDWSCs are extremely separable variants of \tdcs and built by adding an extra layer of separability on \FWSC.  These convolutions are built in three steps (see \figref{3dsc} (right)).  In the first step, similar to \FWSCs, the input cost volume is split into $c_i$ cubes, and $c_i$ disparity-wise separable kernels of size $ k \times k \times 1 $ are applied to each cube to aggregate spatial information and generate $c_i$ output cubes. In the second step, $c_i$ point-wise kernels of size $1\times1\times k$ are applied to each cube independently to aggregate disparity information.  In the final step $c_o$ kernels of size $1\times1\times1\times c_i$ are applied to aggregate information across feature dimension. The total number of operations involved in \FDWSCs is equal to 	$(h_{i} \times w_{i} \times d_{i} \times (k \times k \times 1)) \times c_{i} +  (h_{i} \times w_{i} \times d_{i} \times (1 \times 1 \times k)) \times c_{i}   + ( h_{i} \times w_{i} \times d_{i} \times 1 \times 1 \times 1 \times c_{i} )\times c_{o}$.

As a result, \FDWSCs lead to a further reduction in the number of parameters and operations than all other types of convolutions discussed. Precisely, they lead to a reduction of $3.3\times$ in parameters and  $7.2\times$ in \gmacs for PSMNet and $3.5\times$ in parameters and  $6.0\times$ in \gmacs for \ganetd~--~\cf \tabref{stats-convs}. Surprisingly, this reduction in the number of parameters and operations does not significantly impact the performance -- \cf \tabref{results}.
\begin{figure*}[!t]
\begin{minipage}[b]{0.48\linewidth}
				\renewcommand{\arraystretch}{1.32}
				\vs{0pt}
\resizebox{1\textwidth}{!}{
	\begin{tabular}{|l|l||l|l|l||l|l|l|}
		\hline
\multicolumn{1}{|c}{}&\multicolumn{1}{c}{}&					\multicolumn{3}{|c}{SceneFlow}& 					\multicolumn{3}{|c|}{KITTI 2015}\\
		\hline
		Model                       & Metric        & 3D Convs & \FWSCs            & \FDWSCs                & 3D Convs        & \FWSCs  & \FDWSCs   \\ \hline  \hline 
		\multirow{3}{*}{\ganete}    & $3$ px (\%)   & $4.21$   & $4.02$          & \bm{$3.94$}    & $2.01$          & \bm{$1.94$} & $2.07$\\ \cline{2-8} 
		& D1 error (\%) & $3.49$   & $3.36$          & \bm{$3.25$}  &  $1.92$          & \bm{$1.85$} & $1.95$\\ \cline{2-8} 
		& EPE           & $0.99$   & \bm{$0.92$} & $0.94$     & 			 \bm{$0.67$} & \bm{$0.67$} & $0.68$      \\ \hline \hline
		\multirow{3}{*}{\ganetd} & $3$ px (\%)   & $4.01$   & \bm{$3.74$} & $4.13$  & 				 \bm{$1.67$} & \bm{$1.67$} & $1.73$         \\ \cline{2-8} 
		& D1 error (\%) & $3.29$   & \bm{$3.03$} & $3.45$    & 				 $1.61$          & \bm{$1.57$} & $1.65$       \\ \cline{2-8} 
		& EPE           & $1.01$   & \bm{$0.90$} & $0.99$    & 				\bm{$0.63$} & $0.64$          & $0.71$      \\ \hline  \hline

		\multirow{3}{*}{\psm}     & $3$ px (\%)   & $4.20$   & \bm{$3.72$} & $3.76$  & 				\bm{$2.10$} & $2.57$          & $2.18$        \\ \cline{2-8} 
		& D1 error (\%) & $3.48$   & \bm{$3.08$} & $3.10$     & 					 \bm{$2.00$} & $2.50$          & $2.08$     \\ \cline{2-8} 
		& EPE           & $0.99$   & \bm{$0.90$} & \bm{$0.90$} & 					\bm{$0.88$} & $0.92$          & $0.90$ \\ \hline
	\end{tabular}	
}
	\captionof{table}{Quantitative results of networks trained using \td and \stdcs based \sns on benchmark datasets. }
			\label{table:results}
\end{minipage}\hspace*{2pt}
\begin{minipage}[b]{0.5\linewidth}
\vs{0pt}
\includegraphics[width=0.99\linewidth]{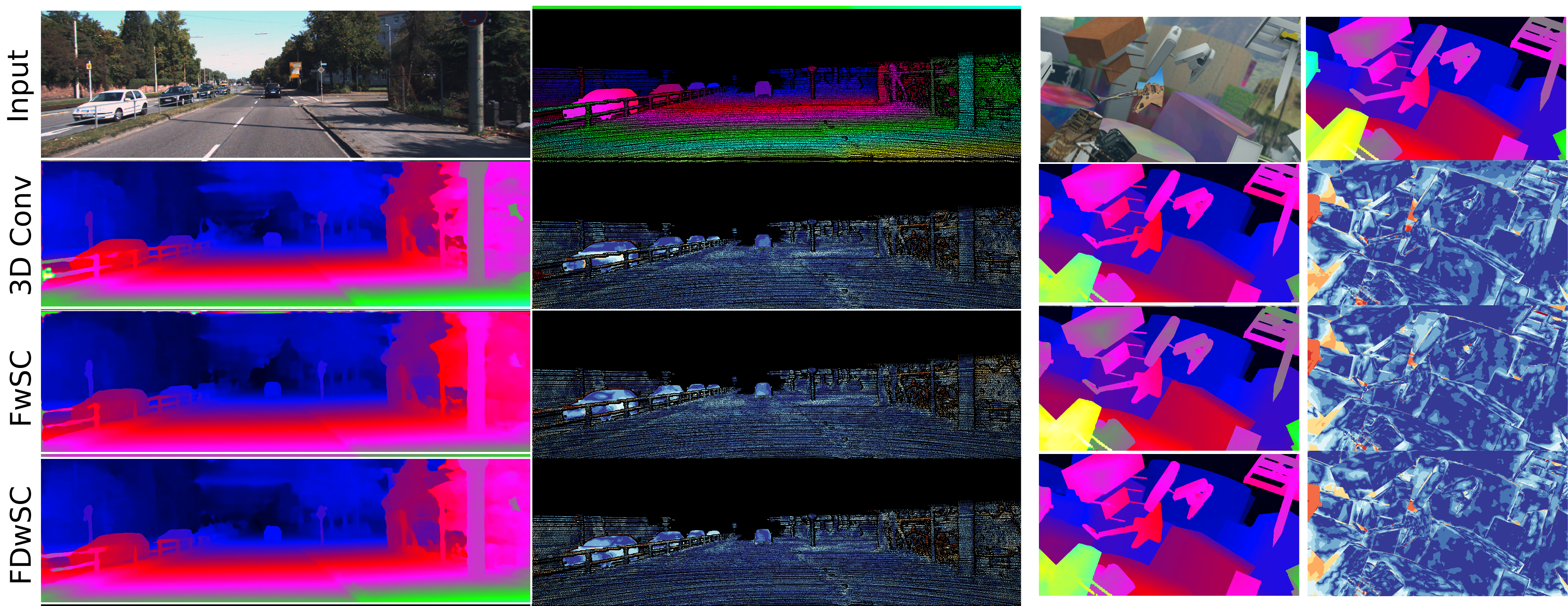}
\captionof{figure}{Qualitative Results of \ganete model on sample images from \kittif (left) and \sf (right) datasets. Here the second and fourth columns represent the error maps \wrt ground truth. Darker red and blue colors represent higher and lower disparity errors, respectively.}
\label{fig:qualitative-results}
\end{minipage}
\vs{-14pt}
\end{figure*}
\vs{-8pt}
\section{Experimental Results}
\label{sec:Results}
\vs{-6pt}
We trained and evaluated our networks on two popular stereo benchmark datasets.\\
\textbf{\sf}~\cite{mayer_large_2016} is a large scale synthetic dataset  and contains $35,454$ training images and $4,370$ test images of size $540 \times 960$ with dense annotations. For our experiments, we further divide the  training dataset to $32, 454$ training and $3000$ validation pairs following  \cite{zhang_ga-net:_2019} .\\
\textbf{\kittif}~\cite{menze_object_2015} is a real dataset of driving scenes. It only contains $200$  training and $200$ test image pairs of size $376 \times 1240$ pixels with sparse annotations. For our experiments, we further divide the training dataset into  $150$ training and $50$ validation pairs as in~\cite{zhang_ga-net:_2019}.
\vs{-12pt}
\subsection{Training Details}
For our baseline networks, we use the publicly released code of \ganetd and \psm. For fair comparison and to show the  ``plug-\&-run" capabilities of \stdcs we use identical training settings for both baseline networks (\ie networks with \tdcs) and optimized networks (\ie networks with \stdcs). Furthermore, it is important to mention that we do not perform any type of hyperparameters tuning and train all the networks from scratch (given in \tabref{results})  with default values as in base networks.\\
\textbf{\ganet:}   All the input images are cropped to the size of $240 \times 528$ while the maximum disparity is set to $192$.  Furthermore, all the images are standard normalized using the mean and standard deviation of the training set. For training, we use the Adam Optimizer ($ \lambda=1\text{e-}3 $, $ \beta_1=0.9  $ and $ \beta_2=0.999 $) with a training batch size of $4$.  For \kittif, we transfer learned the models trained on \sf for $650$ epochs with the learning rate reduced to $1\text{e-}4$ after $ 300 $ epochs.\\
\textbf{\psm:} The majority of settings are identical to the baseline except that the images are randomly cropped to $ 256 \times 512 $ during training, and we use a batch size of $ 8 $. Moreover, for transfer learning on \kittif the learning rate is reduced after $200$ epochs. 
\vs{-12pt}
\subsection{Results}
For quantitative comparison, following the \sf and \kittif evaluation protocols, we report three evaluation metrics including $3$-pixels, D1 and End-Point-Errors (EPE)~\cite{mayer_large_2016,menze_object_2015} for all the networks . On the \sf dataset (\tabref{results} (left)) we notice that for most of the cases, networks with \FWSCs outperform baseline networks with \tdcs despite having around $6.7\times $ less operations and $3.3 \times$  fewer parameters. The better performance of \FWSCs, compared to \tdcs, can be attributed to the fact that a cost volume contains a lot of redundant information (the majority of an image neighbouring disparities do not change) and these \stdcs contain enough representation power to model disparity variations. Surprisingly, even extremely separable \FDWSCs perform better than regular \tdcs in most of the cases. However, their performance is a little inferior to \FWSC, which is somewhat expected, due to the loss of their correlation modeling capabilities in disparity dimension.

By looking at the quantitative results from the \kittif dataset from \tabref{results} (right) we can draw similar conclusions as on the \sf dataset. \FWSCs still outperform \td convolutional networks with a relatively smaller number of \td convolutional layers, \eg \ganetd has $ 15 $ convolutional layers. However, for the networks with a far higher number of \td convolutional layers, \eg \psm ($ 22 $ convolutional layers) we see a slight deterioration in their performance -- this might be because \kittif is a small and sparse dataset.  \FDWSCs also give comparable performance to \FWSCs and \tdcs. 

\figref{qualitative-results} shows qualitative results on sample examples from the \kittif validation and \sf test set. Here once again we can verify that \stdcs are giving comparable results to their \tdcs counterparts.
\vs{-4pt}
\section{Conclusions}
\vs{-6pt}
In this work, we have shown that \td convolutional layers are the major bottleneck in overall execution of \td \sns. To circumvent this we have proposed and empirically shown that \stdcs can considerably reduce the number of operations and parameters for stereo matching in state of the art networks.  These proposed convolutions not only lead to leaner networks but they also consistently lead to better performance of these networks. We believe that these convolutions ``plug-\&-run" nature can lead to their integration  into many other \td \sns.
\label{sec:conclusion}
\newpage
\let\oldbibliography\thebibliography
\renewcommand{\thebibliography}[1]{\oldbibliography{#1}
	\setlength{\itemsep}{-4pt}} 

\bibliographystyle{IEEEbib}
\bibliography{Stereo}

\begin{thebibliography}{10}

\bibitem{scharstein2002taxonomy}
Daniel Scharstein and Richard Szeliski,
\newblock ``A taxonomy and evaluation of dense two-frame stereo correspondence
  algorithms,''
\newblock {\em International journal of computer vision}, vol. 47, no. 1-3, pp.
  7--42, 2002.

\bibitem{hosni2012fast}
Asmaa Hosni, Christoph Rhemann, Michael Bleyer, Carsten Rother, and Margrit
  Gelautz,
\newblock ``Fast cost-volume filtering for visual correspondence and beyond,''
\newblock {\em IEEE TPAMI}, vol. 35, no. 2, pp. 504--511, 2012.

\bibitem{yoon2006adaptive}
Kuk-Jin Yoon and In~So Kweon,
\newblock ``Adaptive support-weight approach for correspondence search,''
\newblock {\em IEEE TPAMI}, vol. 28, no. 4, pp. 650--656, 2006.

\bibitem{min2011revisit}
Dongbo Min, Jiangbo Lu, and Minh~N Do,
\newblock ``A revisit to cost aggregation in stereo matching: How far can we
  reduce its computational redundancy?,''
\newblock in {\em 2011 International Conference on Computer Vision}. IEEE,
  2011, pp. 1567--1574.

\bibitem{zbontar_stereo_2016}
Jure Žbontar and Yann LeCun,
\newblock ``Stereo {Matching} by {Training} a {Convolutional} {Neural}
  {Network} to {Compare} {Image} {Patches},''
\newblock {\em arXiv:1510.05970 [cs]}, May 2016,
\newblock arXiv: 1510.05970.

\bibitem{seki_sgm-nets:_2017}
Akihito Seki and Marc Pollefeys,
\newblock ``{SGM}-{Nets}: {Semi}-{Global} {Matching} with {Neural}
  {Networks},''
\newblock in {\em 2017 {IEEE} {Conference} on {Computer} {Vision} and {Pattern}
  {Recognition} ({CVPR})}, Honolulu, HI, July 2017, pp. 6640--6649, IEEE.

\bibitem{luo2016efficient}
Wenjie Luo, Alexander~G Schwing, and Raquel Urtasun,
\newblock ``Efficient deep learning for stereo matching,''
\newblock in {\em Proceedings of the IEEE CVPR}, 2016, pp. 5695--5703.

\bibitem{kendall2017endtoend}
Alex Kendall, Hayk Martirosyan, Saumitro Dasgupta, Peter Henry, Ryan Kennedy,
  Abraham Bachrach, and Adam Bry,
\newblock ``End-to-end learning of geometry and context for deep stereo
  regression,'' 2017.

\bibitem{chang_pyramid_2018}
Jia-Ren Chang and Yong-Sheng Chen,
\newblock ``Pyramid {Stereo} {Matching} {Network},''
\newblock in {\em 2018 {IEEE}/{CVF} {Conference} on {Computer} {Vision} and
  {Pattern} {Recognition}}, Salt Lake City, UT, June 2018, pp. 5410--5418,
  IEEE.

\bibitem{zhang_ga-net:_2019}
Feihu Zhang, Victor Prisacariu, Ruigang Yang, and Philip H.~S. Torr,
\newblock ``{GA}-{Net}: {Guided} {Aggregation} {Net} for {End}-to-end {Stereo}
  {Matching},''
\newblock {\em arXiv:1904.06587 [cs]}, Apr. 2019,
\newblock arXiv: 1904.06587.

\bibitem{wang_anytime_2019}
Yan Wang, Zihang Lai, Gao Huang, Brian~H. Wang, Laurens van~der Maaten, Mark
  Campbell, and Kilian~Q. Weinberger,
\newblock ``Anytime {Stereo} {Image} {Depth} {Estimation} on {Mobile}
  {Devices},''
\newblock {\em arXiv:1810.11408 [cs]}, Mar. 2019,
\newblock arXiv: 1810.11408.

\bibitem{tulyakov2018practical}
Stepan Tulyakov, Anton Ivanov, and Francois Fleuret,
\newblock ``Practical deep stereo (pds): Toward applications-friendly deep
  stereo matching,'' 2018.

\bibitem{duggal_deeppruner:_2019}
Shivam Duggal, Shenlong Wang, Wei-Chiu Ma, Rui Hu, and Raquel Urtasun,
\newblock ``Deeppruner: {Learning} {Efficient} {Stereo} {Matching} {Via}
  {Differentiable} {Patchmatch},''
\newblock {\em arXiv:1909.05845 [cs]}, Sept. 2019,
\newblock arXiv: 1909.05845.

\bibitem{qiu2017learning}
Zhaofan Qiu, Ting Yao, and Tao Mei,
\newblock ``Learning spatio-temporal representation with pseudo-3d residual
  networks,'' 2017.

\bibitem{ye_3d_2018}
Rongtian Ye, Fangyu Liu, and Liqiang Zhang,
\newblock ``{3D} {Depthwise} {Convolution}: {Reducing} {Model} {Parameters} in
  {3D} {Vision} {Tasks},''
\newblock {\em arXiv:1808.01556 [cs]}, Aug. 2018,
\newblock arXiv: 1808.01556.

\bibitem{poggi_synergies_2020}
Matteo Poggi, Fabio Tosi, Konstantinos Batsos, Philippos Mordohai, and Stefano
  Mattoccia,
\newblock ``On the {Synergies} between {Machine} {Learning} and {Stereo}: a
  {Survey},''
\newblock {\em arXiv:2004.08566 [cs]}, Apr. 2020,
\newblock arXiv: 2004.08566.

\bibitem{mayer_large_2016}
Nikolaus Mayer, Eddy Ilg, Philip Hausser, Philipp Fischer, Daniel Cremers,
  Alexey Dosovitskiy, and Thomas Brox,
\newblock ``A {Large} {Dataset} to {Train} {Convolutional} {Networks} for
  {Disparity}, {Optical} {Flow}, and {Scene} {Flow} {Estimation},''
\newblock in {\em 2016 {IEEE} {Conference} on {Computer} {Vision} and {Pattern}
  {Recognition} ({CVPR})}, Las Vegas, NV, USA, June 2016, pp. 4040--4048, IEEE.

\bibitem{liang_learning_2018}
Zhengfa Liang, Yiliu Feng, Yulan Guo, Hengzhu Liu, Wei Chen, Linbo Qiao,
  Li~Zhou, and Jianfeng Zhang,
\newblock ``Learning for {Disparity} {Estimation} through {Feature}
  {Constancy},''
\newblock {\em arXiv:1712.01039 [cs]}, Mar. 2018,
\newblock arXiv: 1712.01039.

\bibitem{tonioni_real-time_2018}
Alessio Tonioni, Fabio Tosi, Matteo Poggi, Stefano Mattoccia, and Luigi
  Di~Stefano,
\newblock ``Real-time self-adaptive deep stereo,''
\newblock {\em arXiv:1810.05424 [cs]}, Oct. 2018,
\newblock arXiv: 1810.05424.

\bibitem{yang_segstereo_2018}
Guorun Yang, Hengshuang Zhao, Jianping Shi, Zhidong Deng, and Jiaya Jia,
\newblock ``{SegStereo}: {Exploiting} {Semantic} {Information} for {Disparity}
  {Estimation},''
\newblock {\em arXiv:1807.11699 [cs]}, July 2018,
\newblock arXiv: 1807.11699.

\bibitem{song_edgestereo_2018}
Xiao Song, Xu~Zhao, Hanwen Hu, and Liangji Fang,
\newblock ``{EdgeStereo}: {A} {Context} {Integrated} {Residual} {Pyramid}
  {Network} for {Stereo} {Matching},''
\newblock {\em arXiv:1803.05196 [cs]}, Sept. 2018,
\newblock arXiv: 1803.05196.

\bibitem{sandler2018mobilenetv2}
Mark Sandler, Andrew Howard, Menglong Zhu, Andrey Zhmoginov, and Liang-Chieh
  Chen,
\newblock ``Mobilenetv2: Inverted residuals and linear bottlenecks,''
\newblock in {\em Proceedings of the IEEE CVPR}, 2018, pp. 4510--4520.

\bibitem{zhang2018shufflenet}
Xiangyu Zhang, Xinyu Zhou, Mengxiao Lin, and Jian Sun,
\newblock ``Shufflenet: An extremely efficient convolutional neural network for
  mobile devices,''
\newblock in {\em Proceedings of the IEEE CVPR}, 2018, pp. 6848--6856.

\bibitem{menze_object_2015}
Moritz Menze and Andreas Geiger,
\newblock ``Object scene flow for autonomous vehicles,''
\newblock in {\em 2015 {IEEE} {Conference} on {Computer} {Vision} and {Pattern}
  {Recognition} ({CVPR})}, Boston, MA, USA, June 2015, pp. 3061--3070, IEEE.

\end{thebibliography}

\end{document}